\documentclass{llncs}

\usepackage[utf8]{inputenc}
\usepackage[T1]{fontenc}

\usepackage{amsmath, amssymb}
\usepackage{mathtools}
\newcommand{\indep}{\perp \!\!\! \perp}

\usepackage{algorithm}
\usepackage{algorithmicx}
\usepackage[noend]{algpseudocode}

\usepackage{graphicx}
\usepackage{float}

\usepackage{lipsum}

\usepackage{xcolor}
\usepackage{soulutf8}
\usepackage{mdframed}
\mdfsetup{backgroundcolor=yellow}

\usepackage{hyperref}
\hypersetup{
    colorlinks=true,
    linkcolor=blue,
    citecolor=blue,
    filecolor=blue,
    urlcolor=blue
}

\usepackage{cleveref}

\usepackage{footnote}
\makesavenoteenv{tabular}
\makesavenoteenv{table}
\newcommand\blfootnote[1]{%
  \begingroup
  \renewcommand\thefootnote{}\footnote{#1}%
  \addtocounter{footnote}{-1}%
  \endgroup
}

\usepackage[backend=bibtex,style=ieee]{biblatex}
\title{Towards a Transportable Causal Network Model Based on Observational 
Healthcare Data}

\author{
    Alice Bernasconi \inst{1, 2} \orcidID{0000-0001-8522-6882}
    \and Alessio Zanga \inst{1, 3} \orcidID{0000-0003-4423-2121}
\and Peter J.F. Lucas \inst{4} \orcidID{0000-0001-5454-2428}
\and Marco Scutari \inst{5} \orcidID{0000-0002-2151-7266}
    \and Fabio Stella \inst{1} \orcidID{0000-0002-1394-0507}
}

\institute{
    Models and Algorithms for Data \& Text Mining (MADLab), \newline
    University of Milano - Bicocca, Milan, Italy
    \and Evaluative Epidemiology Unit, Department of Epidemiology and Data Science, Fondazione IRCCS Istituto Nazionale dei Tumori, Milan, Italy
    \and Data Science and Advanced Analytics, F. Hoffmann - La Roche Ltd, \newline Basel, Switzerland
    \and University of Twente, Enschede, The Netherlands
    \and Istituto Dalle Molle di Studi sull'Intelligenza Artificiale (IDSIA),
    Lugano, Switzerland
}

\date{\today}

\begin{document}

\maketitle

\begin{abstract}
Over the last decades, many prognostic models based on artificial intelligence techniques have been used to provide detailed predictions in healthcare. Unfortunately, the real-world observational data used to train and validate these models are almost always affected by biases that can strongly impact the outcomes validity: two examples are values missing not-at-random and selection bias. Addressing them is a key element in achieving transportability and in studying the causal relationships that are critical in clinical decision making, going beyond simpler statistical approaches based on probabilistic association.

In this context, we propose a novel approach that combines selection diagrams, missingness graphs, causal discovery and prior knowledge into a single graphical model to estimate the cardiovascular risk of adolescent and young females who survived breast cancer. We learn this model from data comprising two different cohorts of patients. The resulting causal network model is validated by expert clinicians in terms of risk assessment, accuracy and explainability, and provides a prognostic model that outperforms competing machine learning methods.

    \keywords{Causal discovery \and Causal networks \and Transportability \and Missing values \and Selection bias}
\end{abstract}

\section{Introduction}
\label{sec:intro}

An important application of artificial intelligence in healthcare is predicting a disease trajectory conditional on the patient’s history and a projected treatment strategy. In this context, we aim to develop a model that generalizes well: it must provide accurate predictions not only on the \emph{study cohort} (the population from which the data have been collected) but also on the \emph{target cohort}, the general population that it is designed for.  The model's \emph{validity}, accuracy and usefulness in clinical practice are  then of paramount relevance.

In order to establish validity in a rigorous causal framework, we must clearly identify the scope of the study the data come from, as well as the applicability and limitations of the theory we are studying. However, many other issues can impact model validity. Firstly, study cohorts differ from target cohorts in both randomized controlled trials and clinical observational studies due to their inclusion and exclusion criteria. Secondly, \emph{missingness patterns} may introduce bias in the model when the data are \emph{missing not at random} (MNAR). Thirdly, information can be fragmented across different data sets (\emph{data sparsity}) and experts (\emph{domain knowledge}). Therefore, it is important to understand the data selection mechanism and the merging process to ensure the model validity and generalisability \cite{Leung2015}.\\

Our main contributions in this context are:
\begin{itemize}
  \item Developing the first causal network model for estimating the risk of cardiovascular diseases in \emph{adolescent and young adults} that have been treated and survived \emph{breast cancer}.
  \item Developing and adapting a methodological approach to deal with data affected by selection bias and suffering by MNAR.
  \item A thoughtful review of the model we developed and of its implications by domain experts (that is, expert physicians).
\end{itemize}

The rest of the paper is organized as follows. We start by introducing our research question and the topic of the study: \emph{cardio-oncology}, the subfield of cardiology that aims at significantly reducing cardiovascular morbidity and mortality and at improving the quality of life in cancer survivors, here young patients (\Cref{sec:aya}). We complete this introduction by providing an overview of the related work available in the literature (\Cref{sec:related}). We then describe the data (\Cref{sec:data}), the domain knowledge (\Cref{sec:domain-knowledge}) and the causal networks methodology (\Cref{sec:methods}) that we use to design and develop our model. Finally, we describe our findings (\Cref{sec:results}), we contrast them with the available clinical and epidemiological knowledge and we assess model performance against that of other commonly-used machine learning approaches. We complete the paper by summarising our conclusions and presenting future work (\Cref{sec:conclusions}).

\section{Young Adult Breast Cancer Survivors} 
\label{sec:aya}

The increasing number and life expectancy of cancer survivors in the last decades has highlighted the acute and long-term cardio-toxic effects of different cancer therapies. The most common cancer among \emph{adolescent} and \emph{young-adult} females (AYAs; 15--39 years at the first cancer diagnosis) is breast cancer (BC), which is a rare disease with unique genetic and biological features in this age class. AYA female patients are more likely to be genetically susceptible to more aggressive form of BC than older females, and their diagnosis is often delayed because of the lack of BC screening policies for their age group. While lower than in older females with BC, AYAs BC survival rates are high and continuously increasing, thus making such patients likely to be long-term survivors \cite{Trama2016}. Moreover, BC is biologically more aggressive in AYAs than in older women with BC, and requires more aggressive combined neoadjuvant (pre-surgery) and adjuvant (post-surgery) treatments \cite{Schaffar2019}. 

While cardiovascular diseases (CVDs) are well characterized in some groups of cancer survivors, only few studies describe CVDs in AYAs because of their rarity in young patients. Therefore, further evidence is needed to understand and prevent CVDs in AYAs with BC for helping clinicians to plan personalized and effective follow-up guidelines for these patients. Thus, with this work we are interested in answering to the question: \emph{``To what extent is it possible to predict and explain individual susceptibility to cardiotoxicity in AYA with BC?''}, 

\section{Related works} 
\label{sec:related}

Machine learning (ML) techniques have achieved remarkable results in predicting CVDs in cancer patients \cite{Altena2021}. Unfortunately, they are still not part of the clinical practice: cardio-oncologists rely on older, less accurate cardiovascular risk stratification tools such as the Framingham score \cite{Law2017}. The reasons for their reluctance to use ML methods are as follows: 
\begin{itemize} \addtolength{\itemsep}{0.75\baselineskip}
  \item \textbf{Data Availability.} The best-performing ML models such as XGBoost and neural networks rely on biomarkers, laboratory tests, electrocardiograms, echocardiograms, computerized tomography, and cardiac magnetic resonance imaging data \cite{Madan2022}. However, cardio-oncologists rarely have immediate and complete access to such data in their everyday practice.
  \item \textbf{Computational Burden and Data Scarcity.} ML models are known to be computationally intensive and to require large amounts of data to achieve optimal performance. This limits their use in most of healthcare systems, even in rich countries, because healthcare data suffer from quality issues and are typically scarce. They are also often biased and affected by missing-not-at-random patterns \cite{Chen2022} in ways ML models do not account for.
  \item \textbf{High Skills, Knowledge and Expertise.} Current state-of-the-art ML methods require specialized skills, knowledge and expertise to train, validate and deploy especially when combining different types of data, which is almost always the case in healthcare \cite{Nolan2023}. 
  \item \textbf{Lack of Interpretability.} The current state-of-the-art ML models are still difficult to interpret despite recent progress from Explainable AI in addressing this issue \cite{Linardatos2020}.
  \item \textbf{Lack of generalizability.} ML models have been applied successfully to childhood cancer survivors \cite{Chow2015}, where using a limited number of variables, a relatively simple model, and an easy-to-understand user-interface ensured that they were incorporated into clinical practice. However,  cancer is a collection of very complex and heterogeneous diseases, making ML models unlikely to transfer successfully to other cancer survivors cohorts.
  \item \textbf{Correlation vs Causation.} ML models are typically developed using observational data, and in general they can achieve excellent predictive accuracy by leveraging the \emph{association} between the response and the observed variables. However, clinicians operate in a \emph{causal} framework \cite{Bareinboim2016} where they need to evaluate what the outcome could be when prescribing a particular treatment.
\end{itemize}

\section{Materials}

\subsection{Data} 
\label{sec:data}

This project makes use of data coming from two separate cohorts:
\begin{itemize}
  \item \textbf{The population-based cohort (PBC)}: a retrospective cohort of about 1,500 AYAs BC patients who completed cancer treatment, that is, that survived at least 1 year after the cancer diagnosis. In this cohort, BC cases have been identified in population-based cancer registries and the information from each BC patient has been linked to several administrative data sets (hospital discharge records, outpatients and drug flows) for CVD follow-ups \cite{Bernasconi2020}. \\
  \item \textbf{The clinical-based cohort (CBC)}: a retrospective single-institution clinical cohort consisting of 340 additional BC patients, with additional detailed information on cancer prognostic factors but with no CVD follow-up information.\\
\end{itemize}

Around 3\% of AYAs BC patients from the PBC had at least one CVD event during the follow-up (mean follow-up time = 5 years). However, the Framingham score for women with the same characteristics as those in the PBC ranges between 0 and 9, which translates to a predicted risk of 0\% to 1\% of developing a hard coronary heart disease (such as myocardial infraction) or of dying for a CVD event within 10 years \cite{Law2017}. Exploring the PBC further, we found that the most frequent CVDs were due to chemotherapy-induced cardiac damages (arrhythmia and heart failure, in around 1\% of patients) and to ischemic heart diseases that may be related to hormone therapy (in around 2\% of patients). Moreover, 15\% of patients developed at least one major cardiac risk factor (dyslipidemia, diabetes and hypertension) after cancer treatment, thus resulting in an increase in the corresponding Framingham CVD risk. 

The CBC consists of patients treated by a single institution (Fondazione IRCCS Istituto Nazionale dei Tumori di Milano) and is biased due to the patient selection mechanism in ways that are apparent from the baseline and treatment variables distribution. Around 8\% of AYAs BC patients from the CBC have at least one major cardiac risk factor before treatment, compared to only 5\% in the PBC. Moreover, there is an higher proportion of patients receiving neoadjuvant treatments (30\%  in the CBC vs 23\% in the PBC), which are more likely to be prescribed when dealing with later tumor stages at diagnosis (with lymph nodes involvement, metastasis and high dimension tumors). These facts reflect a more severe case-mix of both baseline characteristics of the patients and of tumor aggressiveness in the CBC compared to the PBC, which is expected since the Institute is a referral expertise cancer center.

Furthermore, the PBC and the CBC were collected for different purposes, resulting in different  \emph{missingness mechanisms}. In the CBC, the purpose is patient care, so factors that may influence tumor prognosis (such as tumor grading, staging, etc.) are recorded in greater detail; whereas cancer registry data, which form the basis of the PBC, are collected for administrative, epidemiological monitoring and public health purposes. More details about the distribution of missing values by cohort type is reported in \Cref{tab:missing} in the Supplementary Material.
their percentage of completeness in the cohort of origin.

\subsection{Clinical Knowledge on Cardiovascular Diseases} 
\label{sec:domain-knowledge}

In this section we elicit the clinical knowledge on \emph{cardio-oncology}. To make it easier for the reader to understand how the domain knowledge has been translated from natural language to the structure of the causal network model, we report in square brackets the label of the corresponding nodes. 

The description of the selection mechanism behind the selection node \texttt{[cohort]} has already been described in \Cref{sec:data}.

To decide which treatment to give to a young BC patient, clinicians rely on the most well-known prognostic factors for 5-year cancer survival \texttt{[death\_in\_5y]}  \cite{Cardoso2019}: age (below or above 35 years) \texttt{[age35]}, tumor grade \texttt{[grade]}, tumor histology \texttt{[histology]}, ki67+ status \texttt{[ki67]}, molecular subtype \texttt{[receptors]}, vascular invasion \texttt{[vascular]}, lymph nodes involvement \texttt{[pN]} and tumor dimension \texttt{[pT]}. For example, \emph{triple negative BC} is a specific molecular subtype of BC that is more common in AYAs than in other age groups (prevalence 15\%--20\% \cite{Kim2022}) and whose treatment options are limited to chemotherapy and/or radiotherapy because target therapies are ineffective. 

The risk of CVDs after both neoadjuvant \texttt{[\_neo]} and adjuvant \texttt{[\_adju]} cancer treatments is well known. For instance, anthracycline is frequently prescribed as a chemotherapy regimen \texttt{[chemo\_]}, either alone or in combinations with other chemotherapical drugs. Its cardio-toxic effects are well documented and largely attributable to the generation of free radicals: it ultimately results in left-ventricular dysfunction, arrhythmias \texttt{[cardiotoxicity]} and, in turn, heart failure in the most severe cases \cite{Volkova2012} \texttt{[cvds]}. Even though radiotherapy \texttt{[radio\_]} reduces the risk of cancer recurrence and death in BC patients, the heart may be incidentally exposed to ionizing radiation when the primary cancer is located in the left breast, with in turn increases the risk of heart diseases \texttt{[cvds]} \cite{Taylor2015}. In addition, target therapy \texttt{[target\_]} has been shown to induce acute cardiac toxicity \texttt{[cardiotoxicity]}, especially when administered as an adjuvant treatment (say, trastuzumab for young BC patients  \cite{Mohan2018}), due to possible changes in oxidative stress that reflect mainly in left-ventricular-ejection fraction reduction. However, it is generally believed that the trastuzumab-induced cardiotoxic effects are reversible and that they do not impact long-term CVD risk \cite{An2019}.
 
Moreover, 5 years of tamoxifen \texttt{[hormons\_]}, with or without ovarian suppression/ablation, is considered the standard hormone therapy in young women with hormone-receptor positive disease, which represent the vast majority in this age group. However, it has frequent chronic late effects like type-2 diabetes \texttt{[t2db]}, hypertension \texttt{[hypertension]} and dyslipidemia \texttt{[dyspidemia]} which are all known to be major risk factors for long-term ischemic heart diseases \texttt{[ischemic\_heart\_disease]} \cite{Christinat2013}.

\section{Methods} 
\label{sec:methods}

Assessing validity is a crucial aspect of causal inference. Specifically, \emph{internal validity} refers to \emph{generalizing} the evidence from a study sample to the underlying \emph{study population}; \emph{external validity} refers to \emph{transporting} the evidence to a different \emph{target population}. The interest in the transportability of inference has increased exponentially in the last few years \cite{Degtiar2023ATransportability, Esterling2023TheClaims} thanks to methodological breakthroughs in handling distribution shifts and selection bias. In this work, we leverage both \emph{missingness graphs} \cite{Mohan2018GraphicalData, Tu2018CausalData, Liu2022GreedyValues} and \emph{selection diagrams} \cite{lee2020generalizedb} to learn a causal Bayesian network by combining observational data together with prior knowledge in the presence of selection bias and missingness bias.

\subsection{Causal Models for Missing Values and Selection Bias}

Model-based approaches rely on the formal specification of a \emph{model} representing the interactions we are interested in. In the case of causal inference, \emph{causal graphs} are the de-facto standard for encoding causal relationships.

\begin{definition}[Causal Graph]
  A causal graph $\mathcal{G} = (\mathbf{V}, \mathbf{E})$ \cite{pearl1995causal} is a directed acyclic graph where for each directed edge $(X, Y) \in \mathbf{E}$, $X$ is a direct cause of $Y$ and $Y$ is a direct effect of $X$. The vertex set $\mathbf{V}$ can be split into two disjoint subsets $\mathbf{V} = \mathbf{O} \cup \mathbf{U}$, where $\mathbf{O}$ is the set of the \emph{fully observed} variables (with no missing values) and $\mathbf{U}$ is the set of \emph{fully unobserved} variables, also known as \emph{latent} variables.
\end{definition}

A causal graph allows researches to (i) decide if a consistent estimator exists for a casual effect and (ii) derive that estimator directly from the graph. Being able to correctly estimate causal effects is extremely important for policy making. For instance, it allows clinicians to assess the impact of a given drug on a disease from observational studies when randomized controlled trials are not available for a particular sub-population \cite{Fernandez-Loria2021CausalMattersb}. In more general terms, let $\mathcal{G}$ be a causal graph, $X$ a treatment and $Y$ an outcome. A consistent estimator for the causal effect of $X$ on $Y$ is given by the \emph{do-operator}:
\begin{equation}
  P(Y = y \, | \, do(X = x)) = 
    \sum_\mathbf{z} P(Y = y \, | \, X = x, \mathbf{Z = z})P(\mathbf{Z = z})
\end{equation}
if there exists a set $\mathbf{Z}$ that satisfies the \emph{back-door criterion} for $\mathcal{G}$ \cite{Pearl2009Causality}. If no consistent estimator exists, \emph{confounding bias} makes it impossible to estimate the causal effect correctly. Unbiased estimators of causal effects are theoretically possible even without a causal graph, but their assumptions are rarely satisfied in practical applications \cite{hernan2020whatif}. Therefore, causal graphs remain the tool of choice to achieve both explainability and consistency.

The \emph{missingness mechanism} also plays an important role in causal modeling. Common pre-processing techniques that deal with missing values such as sample deletion and missing imputation are often ineffective or even detrimental to causal effect estimation. For instance, \cite{Stavseth2019HowData, Zanga2023CausalStudy} have shown how missing-data handling has a significant impact on the clinical conclusions that can be drawn from experimental results . Therefore, modeling the reason why a given value is missing is crucial, especially for \emph{MNAR} data. 

The framework for modeling missingness mechanisms is detailed in Rubin's foundational work \cite{RUBIN1976InferenceData}. More recently, \emph{missingness graphs} \cite{Mohan2013GraphicalData, Mohan2018GraphicalData} have been proposed to reconcile Rubin's framework with causal graphs by directly including the missingness indicators in an extended graph structure.

\begin{definition}[Missingness Graph]
  A missingness graph $\mathcal{M} = (\mathbf{V}, \mathbf{E})$ \cite{Mohan2018GraphicalData} is a causal graph whose vertex set $\mathbf{V}$ is partitioned into five disjoint subsets $\mathbf{O} \cup \mathbf{U} \cup \mathbf{M} \cup \mathbf{S} \cup \mathbf{R}$, where: $\mathbf{M}$ is the set of the \emph{partially observed} variables, that is, the variables with at least one missing value; $\mathbf{S}$ is the set of the proxy variables, that is, the variables that are actually observed; $\mathbf{R}$ is the set of the \emph{missingness indicators}.
\end{definition}

Since missingness graphs are extended causal graphs, \emph{d-separation} \cite{koller2009probabilistic} implies (conditional) independence. Briefly, a set of variables $\mathbf{Z}$ d-separates $X$ from $Y$, denoted by $X \indep Y \mid \mathbf{Z}$, if it \emph{blocks} every path  (the combination of all edges and nodes that connect two selected nodes of interest) between $X$ and $Y$. A path is blocked by $\mathbf{Z}$ if and only if it contains: a fork $A \leftarrow B \rightarrow C$ or a chain $A \rightarrow B \rightarrow C$ so that $B$ is in $\mathbf{Z}$, or, a collider $A \rightarrow B \leftarrow C$ so that $B$, or any descendant of it, is not in $\mathbf{Z}$. In this framework, there exists a one-to-one correspondence between the \emph{Missing Completely At Random (MCAR)}, \emph{Missing At Random (MAR)} and \emph{MNAR} patterns and the independence statements implied by the missingness graph:
\begin{itemize}
  \item MCAR implies $\mathbf{O} \cup \mathbf{U} \cup \mathbf{M} \indep \mathbf{R}$: missingness is random and independent from the fully observed variables $\mathbf{O}$ and the partially observed variables $\mathbf{M}$;
  \item MAR implies $\mathbf{U} \cup \mathbf{M} \indep \mathbf{R} \mid \mathbf{O}$: missingness is random only conditionally on the fully observed variables $\mathbf{O}$;
  \item MNAR if neither MCAR nor MAR.
\end{itemize}
This makes it possible to verify whether a consistent estimator for the joint probability $P(\mathbf{X})$ exists in case of MNAR. When the conditions in \cite{Mohan2013GraphicalData} hold, $P(\mathbf{X})$ is \emph{recoverable} and a consistent estimator is given by
\begin{equation}
  P(\mathbf{X}) =
    \frac
      {P(\mathbf{R_X = 0}, \mathbf{X})}
      {
        \prod_{X \in \mathbf{X}}
        P(
            R_X = 0 \, | \,
            \Pi_{R_X},
            \mathbf{R}_{\Pi_{R_X}} = \mathbf{0}
        )
      },
\end{equation}
where $R_X$ is the missingness indicator for variable $X$, $\Pi_{R_X}$ is the parent set of $R_X$ and $\mathbf{R_X}$ is the union of the $R_X$ for all the variables in $\mathbf{X}$.

When the data are a collation of multiple data sets, their different \emph{selection criteria} may induce discrepancies in the distribution of some of the collected variables. In such cases, pooling the data together without modeling the \emph{context} from which observations come from could induce inconsistent estimates \cite{Forre2019CausalBias}. \emph{Selection diagrams} are introduced as an extension of causal graphs for this purpose. We report the definition for completeness.

\begin{definition}[Selection Diagram]
  Let $\Pi$ and $\Pi^*$ be two different populations with a common underlying causal graph $\mathcal{G}$. A selection diagram $\mathcal{S}$ extends the causal graph $\mathcal{G}$ so that:
  \begin{itemize}
    \item $\mathbf{E}_\mathcal{G} \subset \mathbf{E}_\mathcal{S}$, that is, $\mathcal{G}$ is a sub-graph of $\mathcal{S}$;
    \item $\exists \mathbf{S} \neq \varnothing, \mathbf{S} \subset \mathbf{V}_\mathcal{S} \wedge \mathbf{S} \not\subset \mathbf{V}_\mathcal{G}$ where $(S_i \rightarrow V_i) \in \mathbf{E}_\mathcal{S}$ iff $f_i \neq f_i^*$, that is, the variables $\mathbf{S}$ point to the variables $\mathbf{V}_\mathcal{G}$ that differ in their value assignment $f$ in $\Pi$ and $\Pi^*$.
  \end{itemize}
\end{definition}
The variables $\mathbf{S}$ are usually called \emph{selection variables} and allow us to pool together all the observations while keeping track of their provenance. In related work on selection bias, these variables are called \emph{context variables} and identify the context in which $\mathbf{V}_\mathcal{G}$ have been collected. We refer the reader to \cite{Mooij2020JointContexts} for an extended discussion of the differences between selection variables and context variables. In our setting we can use these terms interchangeably. Intuitively, selection bias represents the difference between a consistent estimate for $\Pi$ and a consistent estimate for $\Pi^*$: it limits the our ability to \emph{transport} the inference made on a model for a study population $\Pi$ to another target population $\Pi^*$ due to the differences in their selection criteria. As for d-separation, when a selection diagram $\mathcal{S}$ is modeled, it is possible to identify a consistent estimator for the given set of selection variables $\mathbf{S}$ using the \emph{g-transportability} \cite{Bareinboim2013AResults} criterion.

Once a causal graph $\mathcal{G}$ is designed, we need to connect it to the data $\mathcal{D}$. To do so, we rely on \emph{Casual networks} (CNs) \cite{Pearl1995FromNetworks}, an extension of the well known \emph{Bayesian networks} (BNs) \cite{Pearl2003BAYESIANNETWORKS}, where the underlying graph is a causal graph.

\begin{definition}[Causal Network]
  Let be $\mathcal{G}$ a causal graph and let $P(\mathbf{X})$ be a global probability distribution with parameters $\Theta$. A causal network $\mathcal{C} = (\mathcal{G}, \Theta)$ is a causal model where each variable of $\mathbf{X}$ is a vertex of $\mathcal{G}$ and $P(\mathbf{X})$ factorizes into local probability distributions following:
  \begin{equation}
    P(\mathbf{X}) = \prod_{X \in \mathbf{X}} P(X \, | \, \Pi_X)
  \end{equation}
  where $\Pi_X$ is the parent set of the variable $X$.
\end{definition}

If we combine a missingness graph and a selection diagram into a single causal graph, we can recover from confounding bias, inconsistent estimators due to missing values and domain discrepancies with a single causal network.

\subsection{Causal Discovery with Missing Values and Prior Knowledge}

Manually constructing the \emph{true} causal graph is virtually impossible in real-world applications where there is little to no control over the experimental setting. For instance, there could be unobserved variables affecting the data generating mechanism or unknown interactions between the observed variables. \emph{Causal discovery} \cite{Zanga2022APractice, spirtes2000causation} focuses on recovering the causal graph from collected data and prior knowledge. The existing literature typically assumes that there are no missing values or that the missingness mechanism is either MCAR or MAR \cite{Scutari2020BayesianData}; extensions to MNAR have only appeared in recent years \cite{Tu2018CausalData, Liu2022GreedyValues}.

In this contribution, we take into account the impact of the missingness mechanism using the \emph{Structural Expectation-Maximization} (SEM) algorithm \cite{Lauritzen1995TheData, Friedman1997LearningVariables, Zanga2022RiskApproach} and clinical prior knowledge. At its core, SEM repeats two steps until convergence is reached:
\begin{itemize}
  \item Expectation (E): impute the missing values $\widehat{\mathcal{D}}_i$ from $\widehat{\mathcal{C}}_i$ by estimating $\widehat{\Theta}_i$ using $\widehat{\mathcal{G}}_i$.
  \item Maximization (M): find $\widehat{\mathcal{G}}_{i + 1}$ that maximises a given score function for $\widehat{\mathcal{D}}_i$.
\end{itemize}
At the end of the  $i$th iteration, SEM produces an estimated causal network $\widehat{\mathcal{C}}_{i + 1}$ that will be the starting point of the $(i + 1)$th iteration. In the case of MNAR, it is essential to model the missingness mechanism \emph{a priori} in order to avoid spurious associations induced by the missingness pattern. We achieve that by providing an initial $\mathcal{G}_0$ that encodes a \emph{partially-specified} missingness graph from prior knowledge and by fixing its arcs in place throughout the causal discovery, effectively assuming that they correctly specify the missingness mechanism. Hence, the SEM algorithm only \emph{extends} the initial graph, looking for a super-graph of $\mathcal{G}_0$ that better fits the collected data $\mathcal{D}$.

\subsection{Causal Model Evaluation}

\paragraph{Literature validation.} We validated the edges that are added by the SEM algorithm to the model using the medical literature to support the results relevance in this domain. For this purpose, we performed a literature review to identify publications that explain what we observe in the data.

\paragraph{Performance metrics.} The combined data set comprising the PBC and CBC cohorts was split in a training set and a test set. As mentioned in \Cref{sec:data}, the CBC does not include information on the outcome variables, so we included it only in the training set. The ratio between training and test set was 70\% (60\% PBC + 10\% CBC) / 30\% (PBC). We considered further splitting the training set to obtain a validation set, but that would reduce the sets sample size to the point of making the analysis unfeasible.

We evaluated the CN we learned from the training set on the test set by estimating the probability of CVDs and the associated Area Under the receiver operating characteristic Curve (AUC). In particular, we contrasted the AUC obtained with $\mathcal{G}_0$, with the network learned from SEM (starting from $\mathcal{G}_0$), and with other standard ML methods. AUC represents the degree or measure of separability: it measures how much the model is capable of distinguishing between classes. The higher the AUC, the better the model is at predicting healthy individuals as healthy and individuals affected by CVDs as at risk. 

\section{Experimental Results}
\begin{figure}[H]
  \centering 
  \includegraphics[page=2,width=\linewidth]{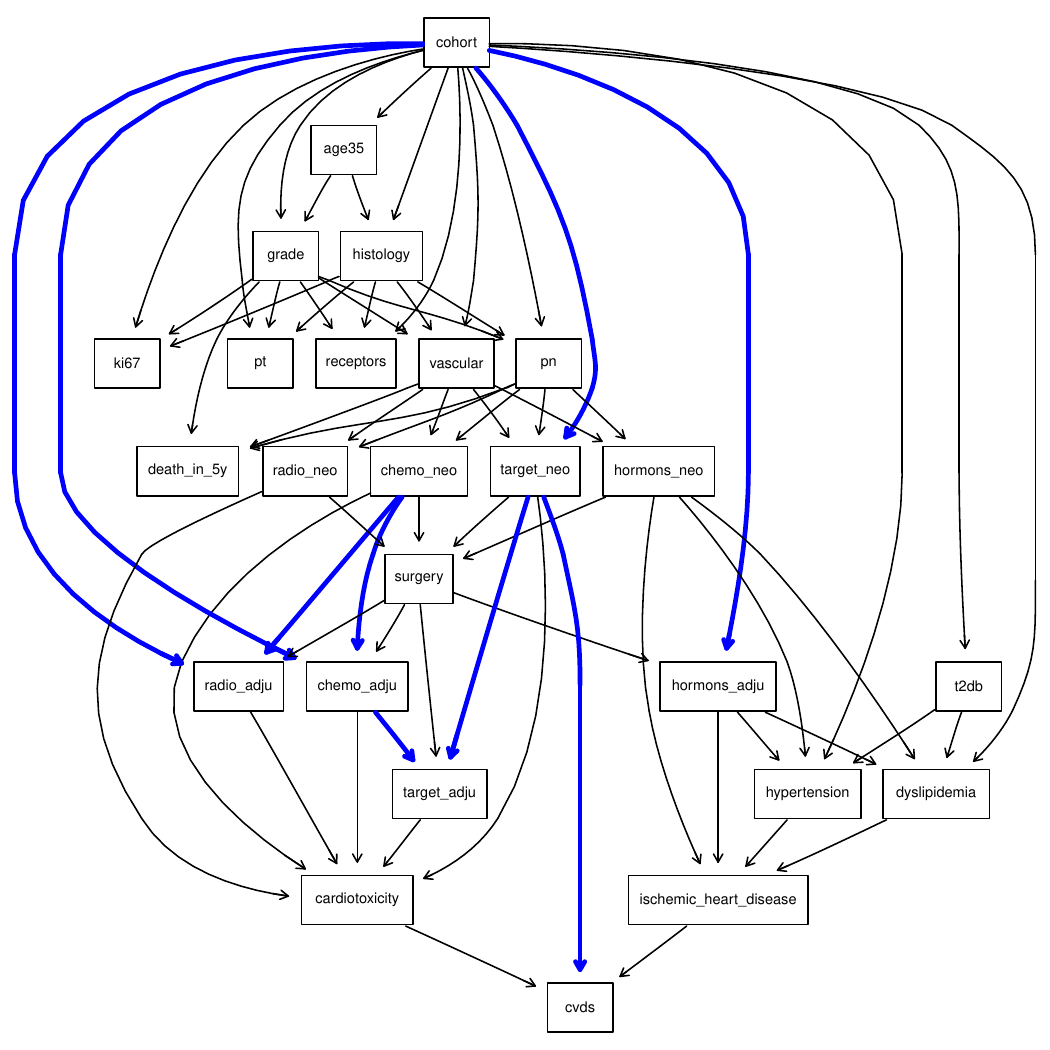}
  \caption{Causal network learned by combining data together with prior knowledge. The edges in black are included in the prior knowledge provided by clinicians, the edges in blue are added by the SEM algorithm.}
  \label{fig:compared-models}
\end{figure}
\label{sec:results}
\paragraph{Validation of SEM algorithm results.} \Cref{fig:compared-models} shows the network learned by SEM: edges elicited from the domain knowledge are in black, while the new ones, that highlight new medical insights, are in \textbf{\textcolor{blue}{bold blue}}. The latter comprise:
\begin{itemize}
  \item \emph{Edges directed from \emph{\texttt{[cohort]}} to the three adjuvant treatments \emph{\texttt{[radio\_adju]}}, \linebreak \emph{\texttt{[chemo\_adju]}}, \emph{\texttt{[hormons\_adju]}} and to one neoadjuvant treatment \emph{\texttt{[target\_neo]}}}. 
  These edges can be explained by the selection mechanism: the treatments nodes act as proxies of case severity \cite{Cardoso2019} which is more likely to be higher in the CBC than in the PBC.\\
  \item \emph{Edges directed from the neoadjuvant treatments \emph{\texttt{[chemo\_neo]}} and \emph{\texttt{[target\_neo]}} to adjuvant treatments \emph{\texttt{[chemo\_adju]}}, \emph{\texttt{[radio\_adju]}} and \emph{\texttt{[target\_adju]}}.}
  These edges can be explained by surgical details not encoded in the \texttt{[surgery]} node, which only describes whether the surgical approach was radical or conservative. The prioritization of breast preservation in AYAs results in more conservative surgical approaches and, in turn, in a higher probability of incomplete surgical margins. As a result, adjuvant treatments are more frequently used to reduce the risk of both local and distant recurrences \cite{Nguyen2021}.\\
  \item \emph{Edges directed from target therapy \emph{\texttt{[target\_neo]}} to cardiovascular diseases \emph{\texttt{[cvds]}}.} 
  These edges can be explained by other unstudied effects of oxidative stress and inflammation that induce CVDs but that are not mediated by cardiotoxicity (for instance, metabolomic and other unmeasured CV risk factors such as obesity) \cite{Jakubiak2021}. Moreover, according to the literature, our understanding of 
 the mechanisms of trastuzumab-mediated cardiotoxicity is still evolving \cite{Mohan2018}. Therefore, they should be further investigated to better tailor the CVDs follow-up guidelines in young BC survivors.
\end{itemize}

\paragraph{Model performance.} The CN model based on the clinical prior knowledge illustrated in \Cref{sec:domain-knowledge} showed a strong classification performance, with an AUC of about  
83\% in the test set. The new edges introduced by the SEM algorithm improved performance to about 88\%. 

It is important to clarify that results obtained from standard ML methods are not directly comparable to those obtained by the proposed approach, for several reasons. Firstly, standard ML methods are not designed to accommodate missing values, relying instead on pre-processing techniques such as sample deletion or imputation. Therefore, only cases coming from the PBC and with complete information on all variables, less than half of the available cases ($\approx$ 800), can be used to train and validate ML models. Secondly, these models do not address the selection bias and lack of external validity. Finally, their results are not causally and clinically interpretable. Against these limitations, we can see in \Cref{tab:model-comparison} that the proposed CN models outperforms all standard ML methods in terms of classification performance.
\blfootnote{\textit{XGBoost: Extreme Gradient Boosting; AODE: Averaged One-Dependence Estimators.}}

\begin{table}[tbh]
\caption{Model performance comparison using Area Under the receiver operating characteristic Curve (AUC).}
\label{tab:model-comparison}
\vspace{0.5em}
\centering
\begin{tabular}{c|c|c|c|c|c|c}
\hline
\multicolumn{1}{c|}{} & 
\multicolumn{1}{|c|}{\textbf{Random}} & 
\multicolumn{1}{|c|}{\textbf{XGBoost}} & 
\multicolumn{1}{|c|}{\textbf{AODE}} & 
\multicolumn{1}{|c|}{\textbf{Naive}}& 
\multicolumn{1}{|c}{\textbf{CN}} &
\multicolumn{1}{|c}{\textbf{CN}} \\
\multicolumn{1}{c|}{} & 
\multicolumn{1}{|c|}{\textbf{Forest}} & 
\multicolumn{1}{|c|}{\textbf{}} & 
\multicolumn{1}{|c|}{\textbf{}} & 
\multicolumn{1}{|c|}{\textbf{Bayes}}& 
\multicolumn{1}{|c|}{\textbf{\emph{prior only}}} &
\multicolumn{1}{|c}{\textbf{\emph{prior}+\emph{\textbf{\textcolor{blue}{SEM}}}}} \\
\hline
\textbf{AUC} & 36.2\% & 65.6\% & 74.0\% & 72.7\% & 82.9\% & 87.6\% \\
 \hline

\end{tabular}
\\
\end{table}

\section{Conclusions and Future work}
\label{sec:conclusions}

The CN shown in \Cref{fig:compared-models} is the first causal network model for estimating the CVD risk in adolescents and young adult females who survived breast cancer. Thanks to the combination of several causal inference methods, including missingness graphs, selection diagrams and causal discovery algorithms, we were able to build a model that combines domain expert knowledge and multiple data sets. The proposed model is able to:
\begin{itemize}
  \item \textit{Deal with the uncertainty and biases} that are intrinsic to real-world observational data and produce valid results, generalizing to other applications or populations;
  \item \textit{Combine data coming from different data sources}, which is especially relevant when dealing with small and understudied populations for which data are extremely sparse.
  \item \textit{Incorporate the domain knowledge} coming from experts, which has been shown to be fundamental in maximizing the model's performance.
  \item \textit{Provide interpretable causal recommendations} to identify those patients that are at higher risk of developing a CVDs and what risk factors are involved, to help physicians in clinical decision making and patient follow-up management.
\end{itemize}

Moreover, the edges added by the causal discovery to the prior-knowledge-based causal network improved its performance and highlighted new arcs that open medical insights particularly valuable, especially for such an understudied population. 

According to our experimental results, the proposed CN model outperforms standard supervised learning methods in terms of classification accuracy: this is unexpected since these methods are specifically trained to optimise it. The are two possible explanations for this phenomenon. Firstly, as mentioned in the results section, the data used to train the models are different: standard ML methods can use only cases coming from the PBC and with complete information on all variables, which means less than half of the cases used to train the CN models. Secondly, the CN models did not learn the model \textit{from scratch} thanks to their ability to incorporate prior domain knowledge; that makes them more efficient at dealing with low sample sizes and missing-not-at-random issues. 

Despite its relevance and strengths, this work has some limitations. First of all, we focused on the classification accuracy of a specific variable of interest instead of evaluating the CN as a whole. This allowed us to compare it with commonly-used ML supervised learning approaches using standard metrics such as AUC, arguing that the proposed CN is competitive in this respect. Having shown that, we are now ready to move beyond prediction into more advanced applications of causal inference which are impossible to carry out with traditional ML models. Moreover, while the arcs added by the SEM algorithm have been validated by a clinical expert and a literature review, a further review by a domain expert (that is, a cardio-oncologist) will be crucial to completely validate the final model. 

In conclusion, we are working on further extending our model to include unmeasured variables, such as distant metastases; this is necessary to further address the need of a clinically relevant model, to help cardio-oncologists in their everyday practice.

\section{Ackwnoledgments}

Alice Bernasconi is funded by an \emph{AIRC} 2020 project, grant number 24864, titled \emph{“pRedicting cardiOvascular diSeAses iN adolescent and young breast caNcer pAtients (ROSANNA)}”.

Alessio Zanga is funded by F. Hoffmann-La Roche Ltd.

This work was partially supported by the MUR under the grant “Dipartimenti di Eccellenza 2023-2027" of the Department of Informatics, Systems and Communication of the University of Milano-Bicocca, Italy.

This work has been submitted and accepted at the \href{https://sites.google.com/unical.it/hcaixia2023/awards-and-accepted-papers}{2nd AIxIA Workshop on Artificial Intelligence For Healthcare (HC@AIxIA 2023)} under \href{https://creativecommons.org/licenses/by/4.0/legalcode.en}{CC BY} License.

\printbibliography

@article{pearl1995causal,
    title = {{Causal diagrams for empirical research}},
    year = {1995},
    journal = {Biometrika},
    author = {Pearl, Judea},
    number = {4},
    pages = {669--688},
    volume = {82},
    publisher = {Oxford University Press},
    doi = {10.1093/biomet/82.4.669},
    issn = {00063444}
}

@book{hernan2020whatif,
    title = {{Causal Inference: What If}},
    year = {2020},
    author = {Hern{\'{a}}n, M A and Robins, J M},
    publisher = {Boca Raton: Chapman {\textbackslash}{\&} Hall/CRC}
}

@book{Pearl2009Causality,
    title = {{Causality: Models, Reasoning and Inference}},
    year = {2009},
    author = {Pearl, Judea},
    edition = {2nd},
    publisher = {Cambridge University Press},
    address = {USA},
    isbn = {052189560X}
}

@book{spirtes2000causation,
    title = {{Causation, prediction, and search}},
    year = {2000},
    author = {Spirtes, Peter and Glymour, Clark N and Scheines, Richard and Heckerman, David},
    publisher = {MIT press}
}

@article{lee2020generalizedb,
    title = {{General Transportability – Synthesizing Observations and Experiments from Heterogeneous Domains}},
    year = {2020},
    journal = {Proceedings of the AAAI Conference on Artificial Intelligence},
    author = {Lee, Sanghack and Correa, Juan and Bareinboim, Elias},
    number = {06},
    month = {4},
    pages = {10210--10217},
    volume = {34},
    url = {https://ojs.aaai.org/index.php/AAAI/article/view/6582},
    doi = {10.1609/aaai.v34i06.6582},
    issn = {2374-3468}
}

@book{koller2009probabilistic,
    title = {{Probabilistic Graphical Models: Principles and Techniques}},
    year = {2009},
    booktitle = {Journal of Chemical Information and Modeling},
    author = {Koller, Daphne and Friedman, Nir},
    pages = {},
    volume = {},
    publisher = {The MIT Press},
    isbn = {978-0-262-01319-2}
}

@article{Bareinboim2013AResults,
    title = {{A General Algorithm for Deciding Transportability of Experimental Results}},
    year = {2013},
    journal = {Journal of Causal Inference},
    author = {Bareinboim, Elias and Pearl, Judea},
    number = {1},
    month = {5},
    pages = {107--134},
    volume = {1},
    url = {https://www.degruyter.com/document/doi/10.1515/jci-2012-0004/html},
    doi = {10.1515/jci-2012-0004},
    issn = {2193-3685}
}

@article{Degtiar2023ATransportability,
    title = {{A Review of Generalizability and Transportability}},
    year = {2023},
    journal = {Annual Review of Statistics and Its Application},
    author = {Degtiar, Irina and Rose, Sherri},
    number = {1},
    month = {3},
    pages = {501--524},
    volume = {10},
    url = {https://www.annualreviews.org/doi/10.1146/annurev-statistics-042522-103837},
    doi = {10.1146/annurev-statistics-042522-103837},
    issn = {2326-8298}
}

@article{Zanga2022APractice,
    title = {{A Survey on Causal Discovery: Theory and Practice}},
    year = {2022},
    journal = {International Journal of Approximate Reasoning},
    author = {Zanga, Alessio and Ozkirimli, Elif and Stella, Fabio},
    number = {},
    month = {12},
    pages = {101--129},
    volume = {151},
    url = {https://linkinghub.elsevier.com/retrieve/pii/S0888613X22001402},
    doi = {10.1016/j.ijar.2022.09.004},
    issn = {0888613X}
}

@article{Scutari2020BayesianData,
    title = {{Bayesian network models for incomplete and dynamic data}},
    year = {2020},
    journal = {Statistica Neerlandica},
    author = {Scutari, Marco},
    number = {3},
    month = {8},
    pages = {397--419},
    volume = {74},
    url = {https://onlinelibrary.wiley.com/doi/10.1111/stan.12197},
    doi = {10.1111/stan.12197},
    issn = {0039-0402}
}

@techreport{Pearl2003BayesianNetworks,
    title = {{Bayesian Networks}},
    year = {2003},
    author = {Pearl, Judea and Russell, Stuart},
    pages = {157--160},
    publisher = {MIT Press}
}

@article{Forre2019CausalBias,
    title = {{Causal calculus in the presence of cycles, latent confounders and selection bias}},
    year = {2019},
    journal = {35th Conference on Uncertainty in Artificial Intelligence, UAI 2019},
    author = {Forr{\'{e}}, Patrick and Mooij, Joris M.},
    arxivId = {1901.00433}
}

@article{Fernandez-Loria2021CausalMattersb,
    title = {{Causal Decision Making and Causal Effect Estimation Are Not the Same... and Why It Matters}},
    year = {2021},
    journal = {the inaugural issue of the INFORMS Journal of Data Science},
    author = {Fern{\'{a}}ndez-Lor{\'{i}}a, Carlos and Provost, Foster},
    month = {4},
    url = {http://arxiv.org/abs/2104.04103},
    arxivId = {2104.04103}
}

@article{Tu2018CausalData,
    title = {{Causal Discovery in the Presence of Missing Data}},
    year = {2018},
    author = {Tu, Ruibo and Zhang, Kun and Ackermann, Paul and Bertilson, Bo Christer and Glymour, Clark and Kjellstr{\"{o}}m, Hedvig and Zhang, Cheng},
    month = {7},
    url = {http://arxiv.org/abs/1807.04010},
    arxivId = {1807.04010}
}

@inproceedings{Zanga2023CausalStudy,
    title = {{Causal Discovery with Missing Data in a Multicentric Clinical Study}},
    year = {2023},
    booktitle = {Proceedings of the 21st International Conference of Artificial Intelligence in Medicine (AIME)},
    author = {Zanga, Alessio and Bernasconi, Alice and Lucas, Peter J. F. and Pijnenborg, Hanny and Reijnen, Casper and Scutari, Marco and Stella, Fabio},
    number = {},
    pages = {40--44},
    volume = {13897 LNAI},
    url = {https://link.springer.com/10.1007/978-3-031-34344-5_5},
    isbn = {9783031343438},
    doi = {10.1007/978-3-031-34344-5{\_}5},
    issn = {16113349}
}

@incollection{Pearl1995FromNetworks,
    title = {{From Bayesian Networks to Causal Networks}},
    year = {1995},
    booktitle = {Mathematical Models for Handling Partial Knowledge in Artificial Intelligence},
    author = {Pearl, Judea},
    pages = {157--182},
    publisher = {Springer US},
    address = {Boston, MA},
    doi = {10.1007/978-1-4899-1424-8{\_}9}
}

@inproceedings{Mohan2013GraphicalData,
    title = {{Graphical models for inference with missing data}},
    year = {2013},
    booktitle = {Advances in Neural Information Processing Systems},
    author = {Mohan, Karthika and Pearl, Judea and Tian, Jin},
    url = {https://proceedings.neurips.cc/paper/2013/file/0ff8033cf9437c213ee13937b1c4c455-Paper.pdf},
    issn = {10495258}
}

@article{Mohan2018GraphicalData,
    title = {{Graphical Models for Processing Missing Data}},
    year = {2018},
    journal = {Journal of the American Statistical Association},
    author = {Mohan, Karthika and Pearl, Judea},
    number = {534},
    month = {1},
    pages = {1023--1037},
    volume = {116},
    url = {https://www.tandfonline.com/doi/full/10.1080/01621459.2021.1874961 http://arxiv.org/abs/1801.03583},
    doi = {10.1080/01621459.2021.1874961},
    issn = {0162-1459},
    arxivId = {1801.03583}
}

@article{Liu2022GreedyValues,
    title = {{Greedy structure learning from data that contain systematic missing values}},
    year = {2022},
    journal = {Machine Learning},
    author = {Liu, Yang and Constantinou, Anthony C.},
    number = {10},
    month = {10},
    pages = {3867--3896},
    volume = {111},
    publisher = {Springer},
    url = {https://link.springer.com/article/10.1007/s10994-022-06195-8},
    doi = {10.1007/S10994-022-06195-8/TABLES/9},
    issn = {15730565},
    arxivId = {2107.04184}
}

@article{Stavseth2019HowData,
    title = {{How handling missing data may impact conclusions: A comparison of six different imputation methods for categorical questionnaire data}},
    year = {2019},
    journal = {SAGE Open Medicine},
    author = {Stavseth, Marianne Riksheim and Clausen, Thomas and R{\o}islien, Jo},
    month = {1},
    pages = {205031211882291},
    volume = {7},
    publisher = {SAGE Publications},
    url = {https://doi.org/10.1177/2050312118822912},
    doi = {10.1177/2050312118822912},
    issn = {2050-3121}
}

@article{Rubin1976InferenceData,
    title = {{Inference and missing data}},
    year = {1976},
    journal = {Biometrika},
    author = {Rubin, Donald D.},
    number = {3},
    pages = {581--592},
    volume = {63},
    url = {https://academic.oup.com/biomet/article-lookup/doi/10.1093/biomet/63.3.581},
    doi = {10.1093/biomet/63.3.581},
    issn = {0006-3444}
}

@article{Mooij2020JointContexts,
    title = {{Joint causal inference from multiple contexts}},
    year = {2020},
    journal = {Journal of Machine Learning Research},
    author = {Mooij, Joris M. and Magliacane, Sara and Claassen, Tom},
    pages = {1--108},
    volume = {21},
    issn = {15337928},
    arxivId = {1611.10351}
}

@inproceedings{Friedman1997LearningVariables,
    title = {{Learning Belief Networks in the Presence of Missing Values and Hidden Variables}},
    year = {1997},
    booktitle = {Proceedings of the Fourteenth International Conference on Machine Learning},
    author = {Friedman, Nir},
    pages = {125--133},
    series = {ICML '97},
    publisher = {Morgan Kaufmann Publishers Inc.},
    address = {San Francisco, CA, USA},
    isbn = {1558604863}
}

@inproceedings{Zanga2022RiskApproach,
    title = {{Risk Assessment of Lymph Node Metastases in Endometrial Cancer Patients: A Causal Approach}},
    year = {2022},
    booktitle = {Proceedings of the 1st Workshop on Artificial Intelligence For Healthcare (HC@AIxIA)},
    author = {Zanga, A. and Bernasconi, A. and Lucas, P.J.F. and Pijnenborg, H. and Reijnen, C. and Scutari, M. and Stella, F.},
    number = {},
    volume = {3307},
    url = {https://www.scopus.com/inward/record.uri?eid=2-s2.0-85145592167&partnerID=40},
    issn = {16130073}
}

@article{Lauritzen1995TheData,
    title = {{The EM algorithm for graphical association models with missing data}},
    year = {1995},
    journal = {Computational Statistics and Data Analysis},
    author = {Lauritzen, Steffen L.},
    number = {2},
    volume = {19},
    doi = {10.1016/0167-9473(93)E0056-A},
    issn = {01679473}
}

@techreport{Esterling2023TheClaims,
    title = {{The Necessity of Construct and External Validity for Generalized Causal Claims}},
    year = {2023},
    author = {Esterling, Kevin M and Brady, David and Schwitzgebel, Eric},
    number = {18},
    publisher = {Institute for Replication (I4R)},
    url = {http://hdl.handle.net/10419/268605},
    address = {s.l.},
    language = {eng}
}

@article{Trama2016,
   author = {Annalisa Trama and Laura Botta and Roberto Foschi and Andrea Ferrari and Charles Stiller and Emmanuel Desandes and Milena Maria Maule and Franco Merletti and Gemma Gatta},
   doi = {10.1016/S1470-2045(16)00162-5},
   issue = {7},
   journal = {The Lancet Oncology},
   pages = {896-906},
   title = {Survival of European adolescents and young adults diagnosed with cancer in 2000–07: population-based data from EUROCARE-5},
   volume = {17},
   year = {2016},
}

@article{Kim2022,
   author = {Hee Jeong Kim and Seonok Kim and Rachel A. Freedman and Ann H. Partridge},
   doi = {10.1016/j.breast.2021.12.006},
   journal = {The Breast},
   pages = {77-83},
   title = {The impact of young age at diagnosis (age \&lt;40 years) on prognosis varies by breast cancer subtype: A U.S. SEER database analysis},
   volume = {61},
   year = {2022},
}

@article{Volkova2012,
   author = {Maria Volkova and Raymond Russell},
   doi = {10.2174/157340311799960645},
   issue = {4},
   journal = {Current Cardiology Reviews},
   pages = {214-220},
   title = {Anthracycline Cardiotoxicity: Prevalence, Pathogenesis and Treatment},
   volume = {7},
   year = {2012},
}

@article{Christinat2013,
   author = {Alexandre Christinat and Simona Di Lascio and Olivia Pagani},
   doi = {10.3978/j.issn.2072-1439.2013.05.25},
   issue = {Suppl 1},
   journal = {Journal of thoracic disease},
   pages = {S36-46},
   title = {Hormonal therapies in young breast cancer patients: when, what and for how long?},
   volume = {5 Suppl 1},
   year = {2013},
}

@article{Law2017,
   author = {W. Law and C. Johnson and M. Rushton and S. Dent},
   doi = {10.3747/co.24.3684},
   issue = {5},
   journal = {Current Oncology},
   pages = {348-353},
   title = {The Framingham Risk Score Underestimates the Risk of Cardiovascular Events in the HER2-Positive Breast Cancer Population},
   volume = {24},
   year = {2017},
}

@article{Altena2021,
   author = {Renske Altena and Laila Hubbert and Narsis A. Kiani and Yvonne Wengström and Jonas Bergh and Elham Hedayati},
   doi = {10.1186/s40959-021-00105-y},
   issue = {1},
   journal = {Cardio-Oncology},
   pages = {20},
   title = {Evidence-based prediction and prevention of cardiovascular morbidity in adults treated for cancer},
   volume = {7},
   year = {2021},
}

@article{Bernasconi2020,
   author = {Alice Bernasconi and Giulio Barigelletti and Andrea Tittarelli and Laura Botta and Gemma Gatta and Giovanna Tagliabue and Paolo Contiero and Stefano Guzzinati and Anita Andreano and Gianfranco Manneschi and Fabio Falcini and Marine Castaing and Rosa Angela Filiberti and Cinzia Gasparotti and Claudia Cirilli and Walter Mazzucco and Lucia Mangone and Silvia Iacovacci and Maria Francesca Vitale and Fabrizio Stracci and Silvano Piffer and Rosario Tumino and Simona Carone and Giuseppe Sampietro and Anna Melcarne and Paola Ballotari and Lorenza Boschetti and Salvatore Pisani and Luca Cavalieri D'Oro and Francesco Cuccaro and Angelo D'Argenzio and Giancarlo D'Orsi and Anna Clara Fanetti and Antonino Ardizzone and Giuseppa Candela and Fabio Savoia and Cristiana Pascucci and Maurizio Castelli and Cinzia Storchi and Annalisa Trama},
   doi = {10.1089/jayao.2019.0170},
   issue = {5},
   journal = {Journal of Adolescent and Young Adult Oncology},
   pages = {586-593},
   title = {Adolescent and Young Adult Cancer Survivors: Design and Characteristics of the First Nationwide Population-Based Cohort in Italy},
   volume = {9},
   year = {2020},
}

@article{Chow2015,
   author = {Eric J. Chow and Yan Chen and Leontien C. Kremer and Norman E. Breslow and Melissa M. Hudson and Gregory T. Armstrong and William L. Border and Elizabeth A.M. Feijen and Daniel M. Green and Lillian R. Meacham and Kathleen A. Meeske and Daniel A. Mulrooney and Kirsten K. Ness and Kevin C. Oeffinger and Charles A. Sklar and Marilyn Stovall and Helena J. van der Pal and Rita E. Weathers and Leslie L. Robison and Yutaka Yasui},
   doi = {10.1200/JCO.2014.56.1373},
   issue = {5},
   journal = {Journal of Clinical Oncology},
   pages = {394-402},
   title = {Individual Prediction of Heart Failure Among Childhood Cancer Survivors},
   volume = {33},
   year = {2015},
}

@article{Bareinboim2016,
   author = {Elias Bareinboim and Judea Pearl},
   doi = {10.1073/pnas.1510507113},
   issue = {27},
   journal = {Proceedings of the National Academy of Sciences of the United States of America},
   pages = {7345-7352},
   title = {Causal inference and the data-fusion problem},
   volume = {113},
   year = {2016},
}

@article{Linardatos2020,
   author = {Pantelis Linardatos and Vasilis Papastefanopoulos and Sotiris Kotsiantis},
   doi = {10.3390/e23010018},
   issue = {1},
   journal = {Entropy},
   pages = {18},
   title = {Explainable AI: A Review of Machine Learning Interpretability Methods},
   volume = {23},
   year = {2020},
}

@article{Nolan2023,
   author = {Paul Nolan},
   doi = {10.1177/00258172221141243},
   journal = {Medico-Legal Journal},
   pages = {002581722211412},
   title = {Artificial intelligence in medicine – is too much transparency a good thing?},
   year = {2023},
}

@article{Chen2022,
   author = {Haidee Chen and David Ouyang and Tina Baykaner and Faizi Jamal and Paul Cheng and June-Wha Rhee},
   doi = {10.3389/fcvm.2022.941148},
   journal = {Frontiers in Cardiovascular Medicine},
   title = {Artificial intelligence applications in cardio-oncology: Leveraging high dimensional cardiovascular data},
   volume = {9},
   year = {2022},
}

@article{Madan2022,
   author = {Nidhi Madan and Julliette Lucas and Nausheen Akhter and Patrick Collier and Feixiong Cheng and Avirup Guha and Lili Zhang and Abhinav Sharma and Abdulaziz Hamid and Imeh Ndiokho and Ethan Wen and Noelle C. Garster and Marielle Scherrer-Crosbie and Sherry-Ann Brown},
   doi = {10.1016/j.ahjo.2022.100126},
   journal = {American Heart Journal Plus: Cardiology Research and Practice},
   pages = {100126},
   title = {Artificial intelligence and imaging: Opportunities in cardio-oncology},
   volume = {15},
   year = {2022},
}

@article{An2019,
   author = {Jie An and M. Saeed Sheikh},
   doi = {10.2174/1568009618666171129222159},
   issue = {5},
   journal = {Current Cancer Drug Targets},
   pages = {400-407},
   title = {Toxicology of Trastuzumab: An Insight into Mechanisms of Cardiotoxicity},
   volume = {19},
   year = {2019},
}

@article{Taylor2015,
   author = {C.W. Taylor and A.M. Kirby},
   doi = {10.1016/j.clon.2015.06.007},
   issue = {11},
   journal = {Clinical Oncology},
   pages = {621-629},
   title = {Cardiac Side-effects From Breast Cancer Radiotherapy},
   volume = {27},
   year = {2015},
}

@article{Schaffar2019,
   author = {Robin Schaffar and Christine Bouchardy and Pierre Olivier Chappuis and Alexandre Bodmer and Simone Benhamou and Elisabetta Rapiti},
   doi = {10.1371/journal.pone.0222136},
   issue = {9},
   journal = {PLOS ONE},
   pages = {e0222136},
   title = {A population-based cohort of young women diagnosed with breast cancer in Geneva, Switzerland},
   volume = {14},
   year = {2019},
}

@article{Jakubiak2021,
   author = {Grzegorz K. Jakubiak and Kamila Osadnik and Mateusz Lejawa and Sławomir Kasperczyk and Tadeusz Osadnik and Natalia Pawlas},
   doi = {10.1155/2021/9987352},
   journal = {Oxidative Medicine and Cellular Longevity},
   pages = {1-19},
   title = {Oxidative Stress in Association with Metabolic Health and Obesity in Young Adults},
   volume = {2021},
   year = {2021},
}

@article{Cardoso2019,
   author = {F. Cardoso and S. Kyriakides and S. Ohno and F. Penault-Llorca and P. Poortmans and I.T. Rubio and S. Zackrisson and E. Senkus},
   doi = {10.1093/annonc/mdz173},
   issue = {8},
   journal = {Annals of Oncology},
   pages = {1194-1220},
   title = {Early breast cancer: ESMO Clinical Practice Guidelines for diagnosis, treatment and follow-up},
   volume = {30},
   year = {2019},
}

@article{Nguyen2021,
   author = {Dang Van Nguyen and Sang-Won Kim and Young-Taek Oh and O Kyu Noh and Yongsik Jung and Mison Chun and Dae Sung Yoon},
   doi = {10.3390/cancers13092150},
   issue = {9},
   journal = {Cancers},
   pages = {2150},
   title = {Local Recurrence in Young Women with Breast Cancer: Breast Conserving Therapy vs. Mastectomy Alone},
   volume = {13},
   year = {2021},
}

@article{Leung2015,
   author = {Lawrence Leung},
   doi = {10.4103/2249-4863.161306},
   issue = {3},
   journal = {Journal of Family Medicine and Primary Care},
   pages = {324},
   title = {Validity, reliability, and generalizability in qualitative research},
   volume = {4},
   year = {2015},
}

@article{Mohan2018,
   author = {Nishant Mohan and Jiangsong Jiang and Milos Dokmanovic and Wen Jin Wu},
   doi = {10.1093/abt/tby003},
   issue = {1},
   journal = {Antibody Therapeutics},
   month = {8},
   pages = {13-17},
   title = {Trastuzumab-mediated cardiotoxicity: current understanding, challenges, and frontiers},
   volume = {1},
   year = {2018},
}

\newpage
\section{Supplementary Material}
\label{sec:supplementary}
\begin{table}[tbh]
  \caption{Percentage of observed values for the variables in the model, categorized according to their role in the domain and stratified by cohort type (PBC, CBC).}
  \label{tab:missing}
\label{tab-variables}
\vspace{0.5em}
\centering
\begin{tabular}{|c|c|c|c|c|}
\hline
\multicolumn{1}{|c|}{\textbf{Variable}}
& \multicolumn{1}{|c|}{\textbf{Label}}
& \multicolumn{1}{|c|}{\textbf{Category}}
& \multicolumn{1}{|c|}{\textbf{PBC}}
& \multicolumn{1}{|c|}{\textbf{CBC}} \\
\hline\hline
Age>35 years & \texttt{age35} & Prognostic factor & 100\% & 100\%\\
Tumor Hystology & \texttt{histology} & Prognostic factor & 100\% & 100\%\\
Tumor grading & \texttt{grade} & Prognostic factor & 0\% & 89\%\\
Vascular invasion & \texttt{vascular} & Prognostic factor & 0\% & 57\%\\
Ki67 positivity & \texttt{ki67} & Prognostic factor & 0\% & 92\%\\
Tumor receptor status & \texttt{receptors} & Prognostic factor & 0\% & 94\%\\
Tumor dimension & \texttt{pT} & Prognostic factor & 0\% & 65\%\\
Lymph nodes involvement & \texttt{pN} & Prognostic factor & 0\% & 68\%\\
Death in 5 years & \texttt{death\_in\_5y} & Survival & 90\% & 45\%\\
Neo-adjuvant chemotherapy & \texttt{chemo\_neo} & Treatment & 100\% & 99\%\\
Neo-adjuvant radiotherapy & \texttt{radio\_neo} & Treatment & 100\%& 99\%\\
Neo-adjuvant target therapy & \texttt{target\_neo} & Treatment & 100\% & 99\%\\
Neo-adjuvant hormon therapy & \texttt{hormons\_neo} & Treatment & 100\% & 99\%\\
Surgery type  & \texttt{surgery} & Treatment & 100\% & 100\%\\
adjuvant chemotherapy & \texttt{chemo\_adju} & Treatment & 100\% & 96\%\\
adjuvant radiotherapy & \texttt{radio\_adju} & Treatment & 100\% & 96\%\\
adjuvant target therapy & \texttt{target\_adju} & Treatment & 100\% & 96\%\\
adjuvant hormon therapy & \texttt{hormons\_adju} & Treatment & 100\% & 96\%\\
Dyslipidemia & \texttt{dyslipidemia} & CVDs risk factor & 100\% & 5\%\\
Hypertension & \texttt{hypertension} & CVDs risk factor & 100\% & 3\%\\
Type 2 Diabete Mellitus & \texttt{t2db} & CVDs risk factor & 100\% & 1\%\\
Cardiotoxicity & \texttt{cardiotoxicity} & CVDs risk factor & 100\% & 0\%\\
Ischemic heart disease & \texttt{ischemic\_heart\_disease} & CVDs risk factor & 100\% & 0\%\\
Cardiovascular disease in 5 years & \texttt{cvds} & Target variable & 100\% & 0\%\\
 \hline
\end{tabular}
\end{table}

\blfootnote{\textit{PBC = Population-based cohort;  CBC = Clinical-based Cohort;  CVDs = Cardiovascular diseases.}}
\end{document}